\documentclass[journal]{IEEEtran}
\usepackage{fixltx2e}
\usepackage{hyperref}
\usepackage{gensymb}
\usepackage{subcaption}
\usepackage{textcomp}

\usepackage{graphicx}
\graphicspath{ {./Figures/} }

\usepackage{epstopdf}
\usepackage{booktabs}

\usepackage{cite}
\ifCLASSINFOpdf
\else
\fi
\usepackage{amsmath}
\usepackage{array}
\begin{document}
% \title{Fully Automatic Myocardial Segmentation of\\ Full-Cycle 2D Contrast Echocardiograms}
\title{Fully Automatic Myocardial Segmentation of Contrast Echocardiography Sequence Using Random Forests Guided by Shape Model }

\author{Yuanwei~Li,
        Chin~Pang~Ho,
        Matthieu~Toulemonde,
        Navtej~Chahal,
        Roxy~Senior,
        and~Meng-Xing~Tang*% <-this % stops a space
\thanks{Y. Li and M. Toulemonde are with the Department
of Bioengineering, Imperial College London, London,
SW7 2AZ UK.}% <-this % stops a space
\thanks{C.P. Ho is with the Department
of Computing, Imperial College London, London,
SW7 2AZ UK.}% <-this % stops a space
\thanks{N. Chahal and R. Senior are with the Department
of Echocardiography, Royal Brompton Hospital, London,
SW3 6NP UK.}% <-this % stops a space
\thanks{*M.X. Tang is with the Department
of Bioengineering, Imperial College London, London,
SW7 2AZ UK (e-mail: mengxing.tang@imperial.ac.uk).}% <-this % stops a space
}


\markboth{IEEE Transactions on Medical Imaging,~Vol.~xx, No.~x, xx~20xx}%
{Li \MakeLowercase{\textit{et al.}}: Myocardial Segmentation Using Random Forests Guided by Shape Model}



\maketitle

\begin{abstract}
Myocardial contrast echocardiography (MCE) is an imaging technique that assesses left ventricle function and myocardial perfusion for the detection of coronary artery diseases. Automatic MCE perfusion quantification is challenging and requires accurate segmentation of the myocardium from noisy and time-varying images. Random forests (RF) have been successfully applied to many medical image segmentation tasks. However, the pixel-wise RF classifier ignores contextual relationships between label outputs of individual pixels. RF which only utilizes local appearance features is also susceptible to data suffering from large intensity variations. In this paper, we demonstrate how to overcome the above limitations of classic RF by presenting a fully automatic segmentation pipeline for myocardial segmentation in full-cycle 2D MCE data. Specifically, a statistical shape model is used to provide shape prior information that guide the RF segmentation in two ways. First, a novel shape model (SM) feature is incorporated into the RF framework to generate a more accurate RF probability map. Second, the shape model is fitted to the RF probability map to refine and constrain the final segmentation to plausible myocardial shapes. We further improve the performance by introducing a bounding box detection algorithm as a preprocessing step in the segmentation pipeline. Our approach on 2D image is further extended to 2D+t sequence which ensures temporal consistency in the resultant sequence segmentations. When evaluated on clinical MCE data, our proposed method achieves notable improvement in segmentation accuracy and outperforms other state-of-the-art methods including the classic RF and its variants, active shape model and image registration.
\end{abstract}


% \fi
\IEEEpeerreviewmaketitle

\section{Introduction} \label{sec:Introduction}
\IEEEPARstart{M}{yocardial} contrast echocardiography (MCE) is a cardiac ultrasound imaging tool that utilizes microbubbles as contrast agents. The microbubbles are injected intravenously and flow within blood vessels. This can improve endocardial visualization and the assessment of left ventricle (LV) structure and function, complementing the conventional B-mode echo \cite{Senior194}. Furthermore, MCE can assess myocardial perfusion through the controlled destruction and replenishment of microbubbles \cite{wei_quantification_1998}. Such perfusion information is useful for the diagnosis of coronary artery diseases (CAD) \cite{Senior194}. 

However, analysis of MCE has been restricted to human visual assessment. Such qualitative assessment is time consuming and relies heavily on the experience of the clinician \cite{DBLP:conf/fimh/MaSRBL09}. Automatic MCE quantification is desired because it is faster, more accurate and less operator-dependent. Quantification can involve the measurements of LV volumes, ejection fraction, myocardial volumes and thickness. It can also involve the assessment of myocardial perfusion by analyzing myocardial intensity changes over time. Myocardial segmentation is a widely used method that serves as an intermediate step to obtaining such MCE quantifications. However, manual segmentation is time-consuming and requires high level of expertise and training. The motivation of this paper is to develop an automatic approach for fast and accurate myocardial segmentation in MCE data. The resultant segmentation can additionally be used as input for subsequent task such as initialization for tracking algorithms \cite{Verhoek2011}.

Although there have been much work on segmentation in B-mode echocardiography, most of these methods do not work well on MCE. MCE utilizes contrast-enhanced ultrasound imaging technique to detect microbubbles by retaining non-linear signals from microbubble oscillations while removing other linear signals from tissue. This is fundamentally different from B-mode echo which simply reflects and captures the linear tissue signals. This results in very different image appearances between MCE and B-mode echo. Specifically, one has to consider the following challenges when performing segmentation on MCE data.
\begin{itemize}
  \item Intensity variations in the image due to ultrasound speckle noise, shadowing and attenuation artefacts, low signal-to-noise ratio, contrast changes over time during microbubble destruction and perfusion imaging \cite{tang_quantitative_2011}.
  \item Geometrical variations in the pose and shape of the myocardium due to different apical chamber views, heart motion and probe motion. Each chamber view is acquired at a different probe position and captures a different 2D cross-section of the LV, resulting in variations of the 2D myocardial shape and orientation.
  \item Misleading intensity information such as 1) presence of structures (papillary muscle) with similar appearance to the myocardium, 2) weak image gradient information resulting in unclear myocardial border (especially epicardium).
  \item Speckle patterns in MCE are decorrelated due to the highly dynamic bubble signals as opposed to the static tissue speckle patterns in B-mode \cite{5441900}. This poses challenges to some tracking algorithms which work by finding corresponding speckle patterns in different frames.
\end{itemize}

\begin{figure*}
\centering
\includegraphics[width=\linewidth]{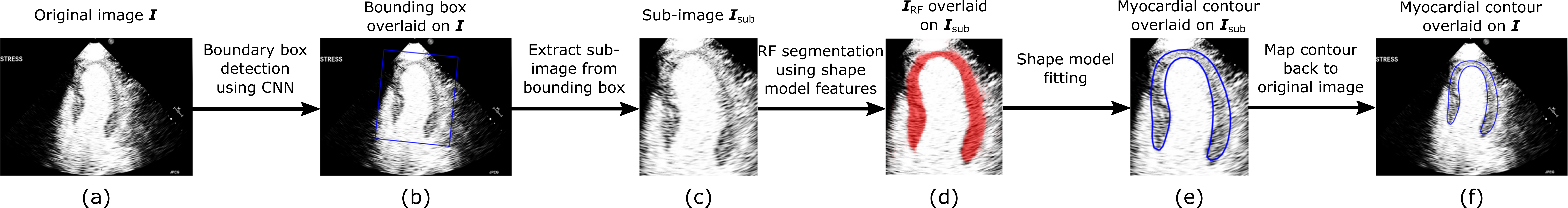}
\caption{Overall pipeline of our myocardial segmentation approach. The three main components of the approach are bounding box detection (a-b), RF segmentation (c-d) and shape model fitting (d-e).}
\label{fig:Pipeline}
\end{figure*}

In this paper, we propose a fully automatic method that segments the myocardium on 2D MCE image. The method is based on random forest (RF) \cite{DBLP:journals/ml/Breiman01} and incorporates global shape information into the RF framework through the use of a statistical shape model that explicitly captures the myocardial shape. This provides stronger and more meaningful structural constraints that guide the RF segmentation more accurately. Our method takes the advantages of both the RF and the shape model in order to address the above challenges of MCE myocardial segmentation. The RF has strong local discriminative power and serves as a good intensity/appearance model that captures the large intensity variations of the MCE data. The shape model captures the myocardial shape variations of the MCE data and imposes a global shape constraint to guide the RF segmentation. This avoids inaccurate segmentation due to misleading intensity information on the MCE. Fig. \ref{fig:Pipeline} shows the overall pipeline of our myocardial segmentation approach. The main contributions of our proposed method are:
\begin{itemize}
  \item A novel shape model (SM) feature is introduced which incorporates the shape model into the RF framework. The feature improves RF segmentation by generating smoother and more coherent RF probability map that conforms to myocardial structure and shape. (Fig. \ref{fig:Pipeline}c-d)
  \item A shape model fitting algorithm is developed to fit the shape model to the RF probability map to produce a smooth and plausible myocardial contour. (Fig. \ref{fig:Pipeline}d-e)
\end{itemize}
The work in this paper is an extension of a preliminary paper \cite{DBLP:conf/miccai/LiHCST16}. The new contributions are:
\begin{itemize}
  \item A convolutional neural network (CNN) is employed to automatically detect a bounding box enclosing the myocardium. This pre-processing step removes any pose variations of the myocardium and improves the subsequent RF segmentation using SM features. The two-stage process of bounding box detection followed by shape inference is similar to \cite{DBLP:journals/mia/Zhou10}. (Fig. \ref{fig:Pipeline}a-b)
  \item The approach is extended to full-cycle 2D+t MCE sequence in which a temporal constraint is imposed to ensure temporal consistency in the sequence segmentations.
  \item The proposed approach is evaluated on a larger dataset of 2D and 2D+t MCE data from 21 subjects.
\end{itemize}

The rest of this paper is structured as follows. In Section \ref{sec:RelatedWork}, we discuss some existing approaches related to cardiac segmentation. In Section \ref{sec:Method}, we describe the different components of our segmentation method and also extends it to sequence segmentation. In Section \ref{sec:Experiments}, we describe the experiments conducted to evaluate the proposed approach. In Section \ref{sec:Results}, we report the segmentation results of our approach and show that it outperforms the other state-of-the-art methods. We conclude the paper in Section \ref{sec:Conclusion} and discuss some limitations and future directions.

\section{Related Work} \label{sec:RelatedWork}
In this section, we give a brief overview of the segmentation methods used for cardiac ultrasound and highlight their limitations when applied to our MCE data. Current literature focus mainly on either LV segmentation \cite{Mishra2003967,Mignotte2001265,Lin2003529,1175086,DBLP:journals/mia/Zhou10,milletari2014left,Oktay2015} or myocardial segmentation \cite{481438,lempitsky_random_2009,7781674,doi:10.1080/21681163.2014.910703,Verhoek2011} in B-mode echocardiography. There are fewer of them addressing segmentation in MCE \cite{DBLP:conf/fimh/MaSRBL09,1710173,malpicaa2004coupled,1625117}. 

Existing methods can be classified into two categories. The first category defines the segmentation task as a contour finding problem in which the optimal contour is found by an optimization procedure based on image information regularized by constraints on the contour. Two representative approaches in this category are the active contour \cite{Kass1988} and active shape model (ASM) \cite{DBLP:journals/cviu/CootesTCG95}. Active contour usually uses a parametric representation of the curves regulalrized by geometric constraints. It is used to detect endocardial border in short-axis B-mode echo by Mishra et al \cite{Mishra2003967} and Mignotte and Meunier \cite{Mignotte2001265}. Malpica et al \cite{malpicaa2004coupled} perform myocardial segmentation in MCE sequence using a coupled active contour which imposes distance constraints between the epi and endocardial contours guided by inter-frame motion estimation derived from optical flow. However, active contour approaches depend on edge information which is unreliable on MCE image. Unclear epicardial border and the presence of papillary muscle often lead to spurious contour estimation. Level sets adopted by Lin et al \cite{Lin2003529} combines edge and region information for LV segmentation. However, region information is adversely affected by the large intensity variation in MCE. A recent paper by Pedrosa et al \cite{7781674} uses the B-spline explicit active surfaces framework for 3D B-mode myocardial segmentation and they compare different ways of coupling epi and endocardial segmentations. 

ASM is another widely used approach in cardiac segmentation. It builds a statistical shape model from a set of training shapes by Principal Component Analysis (PCA). The model is then deform to fit an image in ways similar to the variations observed in the training set. Butakoff et al \cite{doi:10.1080/21681163.2014.910703} uses an automatically constructed ASM for myocardial segmentation in 3D B-mode echo. Pickard et al \cite{1625117} uses ASM for MCE myocardial segmentation and they apply a specialized gradient vector flow field to increase its edged capture range. Bosch et al \cite{1175086} extends ASM by using a joint model of shape, appearance and motion to detect time-continuous LV endocardial contours over time-normalized 2D+t B-mode sequences. The advantage of ASM is that it can account for the myocardial shape variations in our MCE data and provide shape prior to help in cases where the myocardial border is unclear. However, the linear intensity model used by ASM is insufficient to characterize the appearance of MCE data which exhibit huge intensity variations. Therefore, a more powerful non-linear intensity model is required. Furthermore, some of the above methods require good manual initialization of the contour to prevent the solution from getting stuck in poor local minima.

The second category uses machine learning techniques to solve the segmentation task which is often cast as a pixel-wise classification problem. For example, Binder et al \cite{Binder19991069} use an artificial neural network for pixel classification while Xiao et al \cite{981233} combine maximum a posteriori and Markov random field methods for region-based segmentation. For the purpose of this paper, we focus on the RF \cite{DBLP:journals/ml/Breiman01} as the classifier due to its accuracy and computational efficiency. RF has been successfully applied to various segmentation tasks in the medical imaging field \cite{DBLP:journals/neuroimage/GeremiaCMKCA11,DBLP:conf/ipmi/MontilloSWIMC11,DBLP:conf/miccai/MargetaGCA11}. 

RF is well-suited for our MCE data because it effectively builds a non-linear appearance model based on local intensity regions. Therefore, it can better cope with the large intensity variations of the MCE data compared to the simple linear intensity model of ASM. However, the classic RF that utilizes only local appearance features have some limitations. First, the intensity information in MCE may be misleading and result in inaccurate segmentation. For instance, the RF can misclassify the papillary muscle as the myocardium because they have similar intensity. The epicardial boundary is also often difficult to identify based on intensity information alone. Second, the RF classifier assigns a class label to each pixel independently. Structural and contextual relationships between the pixel labels are ignored \cite{DBLP:conf/iccv/KontschiederBBP11,DBLP:conf/ipmi/MontilloSWIMC11}. This results in segmentation with inconsistent pixel labelling which leads to unsmooth boundaries, false detections in the background and holes in the regions of interests. Lastly, RF outputs an intermediate probability map which needs to be post-processed to obtain the final segmentation.

To overcome the above limitations of the classic RF, several works have incorporated local contextual information into the RF framework. Lempitsky et al \cite{lempitsky_random_2009} demonstrates promising myocardial delineation on 3D B-mode echo by using image pixel coordinates as position feature for the RF in order to learn the shape of the myocardium implicitly. Verhoek et al \cite{Verhoek2011} further extends the method by using optical flow to propagate the single-frame RF segmentation for sequence tracking. Montillo et al \cite{DBLP:conf/ipmi/MontilloSWIMC11} introduces the entangled RF which uses intermediate probability predictions from higher levels of the tree as features for training the deeper levels. Kontschieder et al \cite{DBLP:conf/iccv/KontschiederBBP11} introduces the structured RF that incorporates structural relationships into the output predictions by predicting structured class labels for a patch region rather than an independent class label for each individual pixel. This is used by Oktay et al \cite{Oktay2015} to extract a new boundary representation of the 3D B-mode echo data which is then utilized in a multi-atlas framework for LV segmentation. Milletari et al \cite{milletari2014left} employs a Hough forests with implicit shape and appearance priors for the simultaneous detection and segmentation of LV in 3D B-mode echo. The Hough forest votes for the location of the LV centroid and the contour is subsequently estimated by using a code-book of surface patches associated with the votes. All the above works use contextual information to improve the RF segmentation. But they only impose weak structural constraint locally and do not explicitly learn the myocardial shapes from our MCE data.

\section{Method} \label{sec:Method}
The overall pipeline for our myocardial segmentation method is summarized in Fig. \ref{fig:Pipeline}. Given an input MCE image $\boldsymbol{I}$, a bounding box enclosing the myocardium is first detected using a CNN. Using the bounding box, a sub-image $\boldsymbol{I}_{\textrm{sub}}$ is cropped out from $\boldsymbol{I}$ and then rescaled to a fixed size. Next, an RF classifier with SM feature is used to predict a myocardial probability map $\boldsymbol{I}_{\textrm{RF}}$ of the sub-image. A statistical shape model is then fitted to $\boldsymbol{I}_{\textrm{RF}}$ to give a final myocardial contour which is subsequently mapped back to the original image space. For sequence segmentation, an additional constraint term is added to the shape model fitting step to ensure temporal consistency in the segmentations.

%\IEEEpubidadjcol

\subsection{Bounding Box Detection} \label{sec:MethodBB}
As we will show later (Section \ref{sec:ResultsBB}), our SM features only work well on RF input images which do not contain significant pose variations (translation, scaling, rotation) in the myocardium. This requires the rigid alignment of myocardium in RF input images. This can be done by image registration \cite{DBLP:journals/tmi/RueckertSHHLH99} but the running time can be slow. To this end, we employ a CNN to automatically detect a bounding box containing the myocardium (Fig. \ref{fig:Pipeline}b). The bounding box is then used to extract a sub-image from the original image and the sub-image is rescaled to a fixed size. This ensures the RF input image $\boldsymbol{I}_{\textrm{sub}}$ is free from pose variations in the myocardium. CNN has been proven to be a good method in many object detection tasks \cite{DBLP:journals/corr/SermanetEZMFL13,DBLP:conf/cvpr/GirshickDDM14,DBLP:journals/pami/GarciaD04} and it is also computationally efficient when implemented on GPUs. In our case, CNN can automatically learn a hierarchy of low to high level features to predict the myocardial pose in an MCE image accurately.

\def\arraystretch{1.2}
\begin{table}[]
\centering
\caption{CNN architecture for bounding box detection. C and FC denote convolutional layer and fully-connected layer respectively. Kernel parameters are given by [kernel height $\times$ kernel width $\times$ number of kernels / stride]. }
\label{table:MethodCNN}
\begin{tabular}{m{1cm} m{2.5cm} m{1cm} m{2.5cm} l}
\hline \hline
Layer & Kernel parameters & Layer & Kernel parameters \\
\hline
1. C1 & $[11\times 11\times 96$ / $4]$ & 7. \space C5 & $[3\times 3\times 256$ / $1]$ \\
\multicolumn{2}{l}{2. Max pooling and LRN} & \multicolumn{2}{l}{8. \space Max pooling} \\
3. C2 & $[5\times 5\times 256$ / $1]$ & 9. \space FC6 & $[1\times 1\times 4096$ / $1]$ \\
\multicolumn{2}{l}{4. Max pooling and LRN} & 10. FC7 & $[1\times 1\times 4096$ / $1]$ \\
5. C3 & $[3\times 3\times 384$ / $1]$ & 11. FC8 & $[1\times 1\times 5$ / $1]$ \\
6. C4 & $[3\times 3\times 384$ / $1]$ \\
\hline \hline
\end{tabular}
\end{table}

Similar to \cite{DBLP:journals/mia/Zhou10}, we define the bounding box $\boldsymbol{B}$ using 5 parameters $(B_x,B_y,B_w,B_h,B_\theta)$ which represent its centroid $(B_x,B_y)$, size $(B_w,B_h)$ and orientation $B_\theta$ respectively. Bounding box detection is cast as a regression problem using CNN in which the values of the bounding box parameters are predicted. We use the CaffeNet architecture which is a slight modification of the AlexNet \cite{DBLP:conf/nips/KrizhevskySH12}. Table \ref{table:MethodCNN} shows the network architecture comprising 5 convolutional layers and 3 fully-connected layers together with the kernel parameters for each layer. Max pooling is always performed using a $3\times3$ kernel with stride=2. Rectified linear units (ReLU) is the activation function applied after all convolutional layers. Local response normalisation (LRN) is performed after the max pooling operations at layer 2 and 4. Dropout layer is added after the fully-connected layers, FC6 and FC7. The final classification layer in the original architecture is replaced with a regression layer FC8 which contains 5 units that correspond to the 5 bounding box parameters. The network is trained using $l_2$ Euclidean loss between the predicted and ground truth bounding box parameters. During training, the original $460\times 643$ image $\boldsymbol{I}$ is downsampled to a size of $256\times 358$ pixels. Random cropping ($227\times 227$ pixels) and rotation (within $\pm 10\degree$) are then performed to prepare the training samples which are presented to the network in mini-batches of 32. The network is initialized with random weights from $(\mu, \sigma )=(0, 0.01)$. Optimization is performed using Adam algorithm with learning rate=${ 10 }^{ -4 }$, momentum=0.9 and weight decay=${ 5\times 10 }^{ -4 }$. The training is run for 18000 iterations. During testing, we extract 55 crops at uniform intervals from the downsampled test image and combine their predictions to give the final predicted bounding box. A sub-image is then extracted using this bounding box and it is rescaled to a size of $242\times 208$ pixels to be used as input to the RF.

\subsection{Shape Model Construction} \label{sec:MethodModel}
\begin{figure}
\centering
\begin{subfigure}[b]{.116\textwidth}
  \centering
  \includegraphics[width=\linewidth]{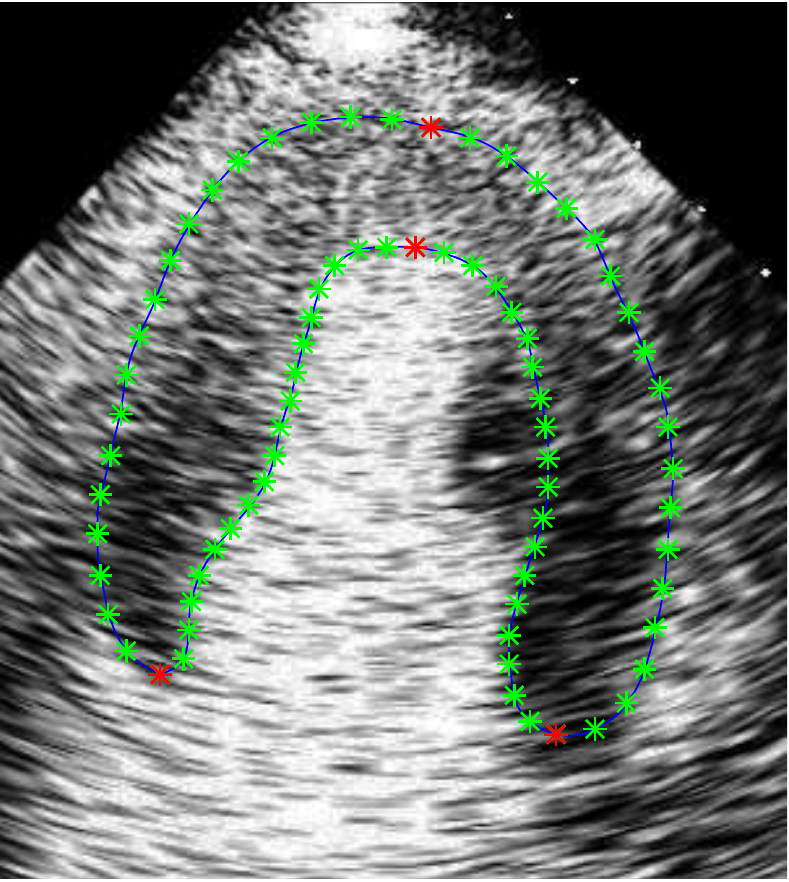}
  \caption{}
  \label{fig:TrainingShapes}
\end{subfigure}%
\begin{subfigure}[b]{.152\textwidth}
  \centering
  \includegraphics[width=\linewidth]{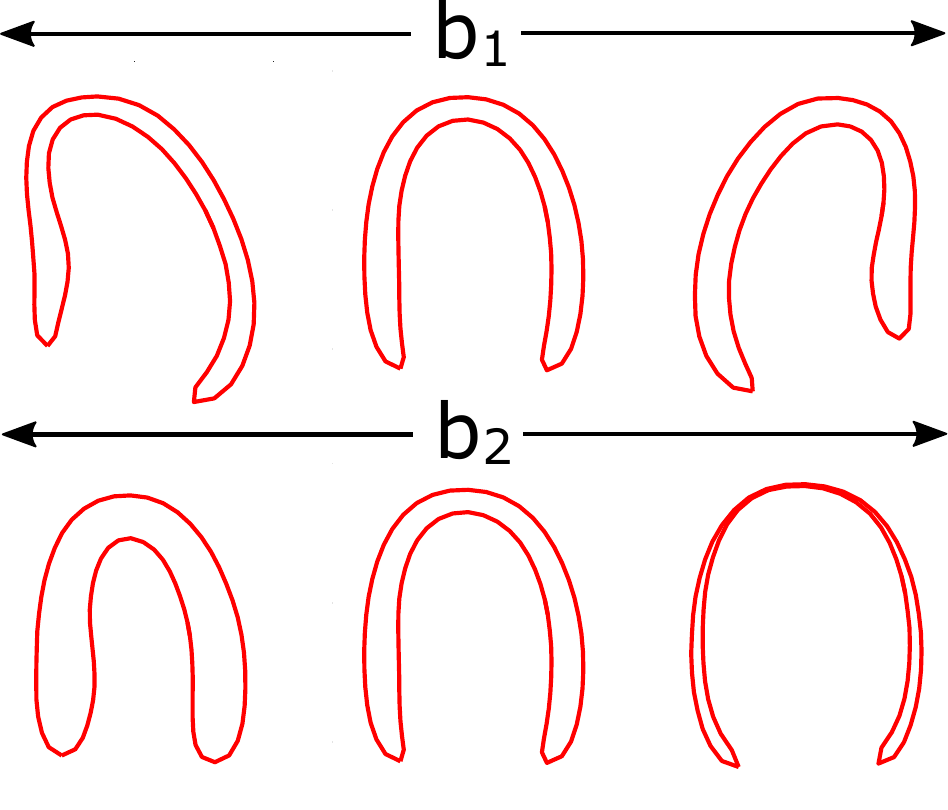}
  \caption{}
  \label{fig:ShapeModel}
\end{subfigure}%
\begin{subfigure}[b]{.232\textwidth}
  \centering
  \includegraphics[width=1\linewidth]{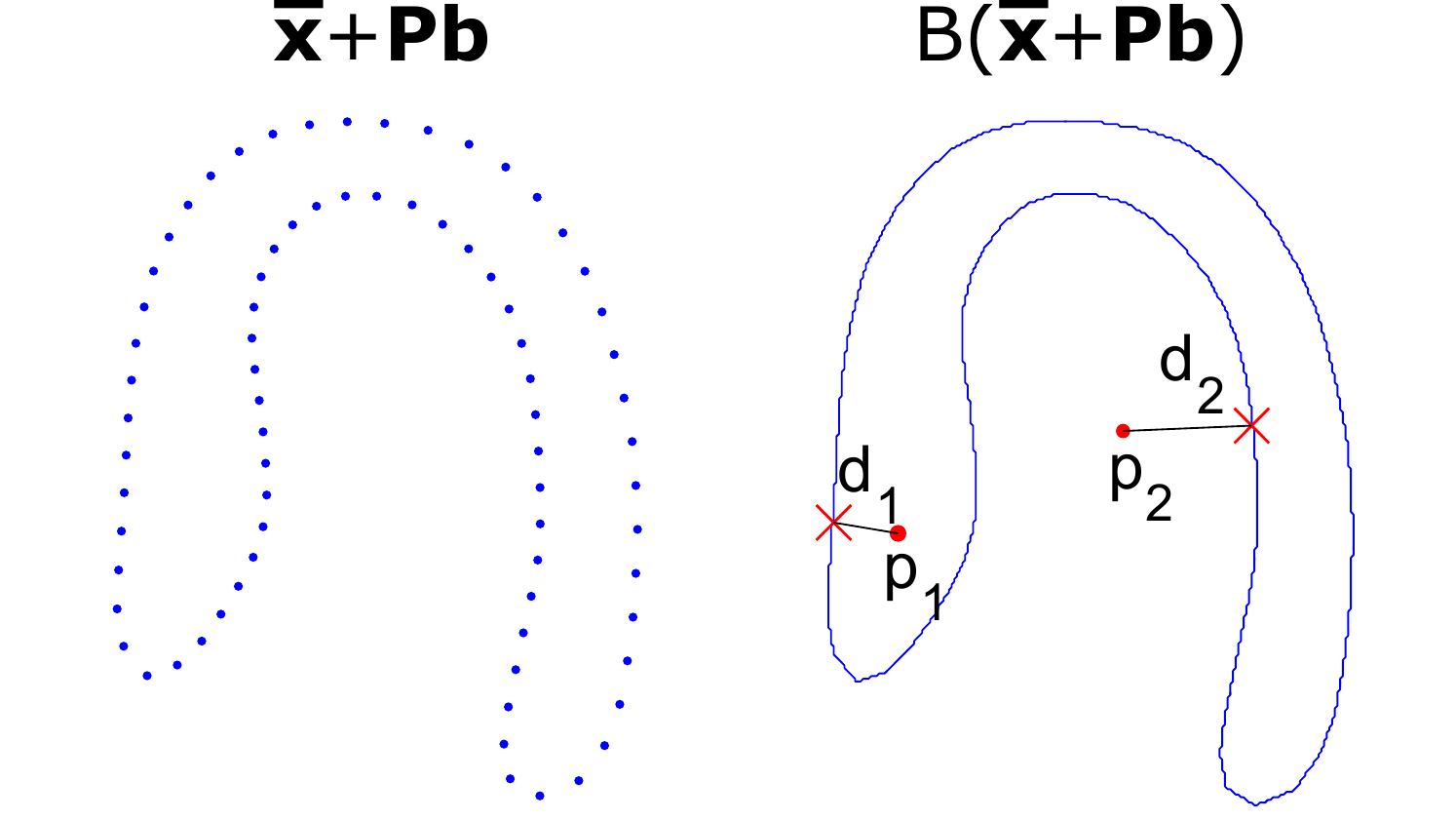}
  \caption{}
  \label{fig:SMFeature}
\end{subfigure}
\caption{(a) Manual annotations showing key landmarks in red and other landmarks in green. (b) First and second modes of variation of the shape model with $b_i$ varying in the range of $\pm 2\sqrt { { \lambda  }_{ i } } $. (c) SM feature computation. Left: Landmarks $\boldsymbol{x}$ generated randomly by the shape model in (\ref{eq:Model}) (\textit{blue dots}). Right: SM feature values $d_1$ and $d_2$ measure the signed shortest distance from boundary $B$ (\textit{blue contour}) to pixel $\boldsymbol{p_1}$ and $\boldsymbol{p_2}$ respectively. $d_1$ is positive and $d_2$ is negative.}
\label{fig:Result}
\end{figure}

In this section, we explain the construction of a statistical shape model of the myocardium which is required for the subsequent stages of our segmentation pipeline. Statistical shape model is extremely useful for representing objects with complex shapes. It captures plausible shape variations while removing variations due to noise. The 2D myocardial shape can be effectively represented using the point distribution model \cite{DBLP:journals/cviu/CootesTCG95} which uses a set of landmarks to describe the shape. The model is built from a set of training shapes using PCA which captures the correlations of the landmarks among the training set. Each training shape is defined by a set of $M$ landmarks. It consists of 4 key landmarks with the other landmarks uniformly sampled in between (Fig. \ref{fig:TrainingShapes}). The key landmarks are the two apexes on the epicardium and endocardium and the two endpoints on the basal segments. The point distribution model is given by:
\begin{equation}\label{eq:Model}
  \boldsymbol{x}=\bar { \boldsymbol{x} } + \boldsymbol{Pb}
\end{equation}
where $\boldsymbol{x}$ is a 2$M$-dimensional vector $(x_1,y_1,...,x_M,y_M)$ containing the $x$,$y$-coordinates of the $M$ landmarks, $\bar { \boldsymbol{x} }$ is the mean landmark coordinates of the training shapes, $\boldsymbol{P}=({ \boldsymbol{p} }_{ 1 }|{ \boldsymbol{p} }_{ 2 }|...|{ \boldsymbol{p} }_{ K })$ contains $K$ eigenvectors of the covariance matrix and each $\boldsymbol{p}_i$ is associated with its eigenvalue $\lambda_i$, $\boldsymbol{b}$ is a $K$-dimensional vector containing the shape parameters where each element $b_i$ is bounded between $\pm s\sqrt { { \lambda  }_{ i } } $ to ensure that only plausible myocardial shapes are produced. $s$ is the number of standard deviation from the mean shape. The value of $K$ can be chosen such that the model can explain a required percentage $p$ of the total variance present in the training shapes. The shape model is built from manual annotations represented in the cropped coordinate space of $\boldsymbol{I}_{\textrm{sub}}$. Since images extracted from bounding box are already corrected for pose variations, we do not need to rigidly align the training shapes prior to PCA. Fig. \ref{fig:ShapeModel} shows the first and second modes of variation of our myocardial shape model.

%\begin{figure}
%\centering
%\begin{subfigure}[b]{.29\textwidth}
%  \centering
%  \includegraphics[width=\linewidth]{TrainingShapes.pdf}
%  %\includegraphics[height=.2\textheight]{Fig3}
%  \caption{}
%  \label{fig:TrainingShapes}
%\end{subfigure}%
%\begin{subfigure}[b]{.19\textwidth}
%  \centering
%  \includegraphics[width=\linewidth]{ShapeModelFinal.pdf}
%  %\includegraphics[height=.2\textheight]{Fig4_matlab_bigFont}
%  \caption{}
%  \label{fig:ShapeModel}
%\end{subfigure}
%\caption{(a) Examples of manual landmark annotations showing the key landmarks in red and the other landmarks in green. (b) First and second modes of variation of the myocardial shape model with $b_i$ varying within the range of $\pm 2\sqrt { { \lambda  }_{ i } } $.}
%\label{fig:Result}
%\end{figure}

%\begin{table*}[]
%\centering
%\caption{Parameters of constructed shape models.}
%\label{table:SMParams}
%\begin{tabular}{c c c c c}
%\hline \hline
%Shape models & $Model_{RF}$ & $Model_{Fit}$ & $Model_{Vid1}$ & $Model_{Vid2}$ \\
%\hline
%K & 14 & 35 & Variable & Variable \\
%Total variance explained & 98\% & 99.9\% & 98\% & 99.9\% \\
%s & 1 & 2 & - & 2 \\
%Training dataset & $Dataset1$ & $Dataset1$ & All frames in video & Non-outlier frames in video \\
%\hline \hline
%\end{tabular}
%\end{table*}

%\def\arraystretch{1.2}
%\begin{table}[]
%\centering
%\caption{Parameters and training datasets used for the construction of different shape models.}
%\label{table:SMParams}
%\begin{tabular}{m{1.33cm} m{1.33cm} m{1.33cm} m{1.33cm} m{1.33cm}}
%\hline \hline
%Shape models & $Model_{RF}$ & $Model_{Fit}$ & $Model_{Vid1}$ & $Model_{Vid2}$ \\
%\hline
%K & 14 & 35 & Variable & Variable \\
%Total variance explained & 98\% & 99.9\% & 98\% & 99.9\% \\
%s & 1 & 2 & - & 2 \\
%Training dataset & $Dataset1$ & $Dataset1$ & All frames in video & Non-outlier frames in video \\
%\hline \hline
%\end{tabular}
%\end{table}

%\begin{figure}
%\centering
%\includegraphics[width=\linewidth]{TrainingShapes.pdf}
%\caption{Examples of manual landmark annotations showing key landmarks in red and the other landmarks in green.}
%\label{fig:TrainingShapes}
%\end{figure}

\subsection{RF with SM Feature} \label{sec:MethodFeature}
Myocardial segmentation using RF can be treated as a pixel-wise binary classification problem \cite{lempitsky_random_2009}. The RF takes the sub-image $\boldsymbol{I}_{\textrm{sub}}$ from Section \ref{sec:MethodBB} as input and predicts the class label (myocardium or background) of every pixel in that sub-image based on a set of features. An RF is an ensemble of decision trees. During training, each split node of the tree learns a binary feature test that best splits the training pixels into its two child nodes by maximizing the information gain. The splitting continues recursively until the maximum tree depth is reached or the number of training pixels falls below a minimum. At this time, a leaf node is created and the class label distributions of the training pixels reaching that node is stored and used to predict the class probability of future unseen test pixels. During testing, the predictions from all the trees are averaged to give a final myocardial probability map.

In the classic RF, the binary feature test is based on appearance/intensities of local image regions. This is either the mean pixel intensity of a displaced feature box from the reference pixel or the difference of the mean pixel intensities of two such feature boxes \cite{DBLP:journals/mia/CriminisiRKSPWS13}. We introduce an additional novel SM feature \cite{DBLP:conf/miccai/LiHCST16} that is derived from the shape model constructed in Section \ref{sec:MethodModel}). An SM feature is constructed by randomly sampling some values of $\boldsymbol{b}$ to generate a set of landmarks $\boldsymbol{x}$ using (\ref{eq:Model}) (Fig. \ref{fig:SMFeature} left). A closed myocardial boundary $B$ is then formed from the landmarks through linear interpolation. The SM feature is then given by the signed shortest distance $d$ from the reference pixel $\boldsymbol{p}$ to the myocardial boundary $B$ (Fig. \ref{fig:SMFeature} right). $d$ is positive if $\boldsymbol{p}$ lies inside $B$ and negative otherwise. The SM feature is in fact a signed distance transform of the myocardial boundary generated by the shape model. Each SM feature is defined by the shape parameters $\boldsymbol{b}$. During training, an SM feature is constructed by random uniform sampling of each $b_i$ values in the range of $\pm s\sqrt { { \lambda  }_{ i } }$. The binary SM feature test, parameterized by $\boldsymbol{b}$ and a threshold $\tau$, is written as:
\begin{equation}
{ t }_{ \textrm{SM} }^{ \boldsymbol{b},\tau  }(\boldsymbol{p})=\begin{cases} 1,\qquad \textrm{if}\quad D(\boldsymbol{p},B(\bar { \boldsymbol{x} } +\boldsymbol{Pb}))>\tau  \\ 0,\qquad \textrm{otherwise}. \end{cases}
\end{equation}
where $D(.)$ is the function that computes $d$, $B$ is the function that converts the landmarks to a closed boundary by linear interpolation. Depending on the binary test outcome, pixel $\boldsymbol{p}$ will either go to the left (1) or right (0) child node. During training, the RF learns the values of $\boldsymbol{b}$ and $\tau$ that best split the training pixels at the split node. The SM features explicitly impose a global shape constraint in the RF framework. The random sampling of $\boldsymbol{b}$ also allows the RF to learn plausible shape variations of the myocardium.

%\begin{figure}
%\centering
%\includegraphics[width=0.7\linewidth]{SM_feature.pdf}
%\caption{Left: Landmarks $\boldsymbol{x}$ are generated randomly by the shape model in (\ref{eq:Model}) after which a random pose transformation $T_{\boldsymbol{\theta}}$ is applied (\textit{blue dots}). Right: $d_1$($d_2$) is the SM feature value measuring the signed shortest distance from pixel $\boldsymbol{p_1}$($\boldsymbol{p_2}$) to the myocardial boundary $B$ (\textit{blue contour}). $d_1$ is positive and $d_2$ is negative.}
%\label{fig:SMFeature}
%\end{figure}

%\begin{figure}
%\centering
%\begin{subfigure}[b]{0.2\textwidth}
%  \centering
%  \includegraphics[width=0.9\linewidth]{SM_feature_a.pdf}
%  %\includegraphics[height=.2\textheight]{Fig3}
%  \caption{}
%  \label{fig:SMFeature_a}
%\end{subfigure}%
%\begin{subfigure}[b]{0.2\textwidth}
%  \centering
%  \includegraphics[width=0.9\linewidth]{SM_feature_b.pdf}
%  %\includegraphics[height=.2\textheight]{Fig4_matlab_bigFont}
%  \caption{}
%  \label{fig:SMFeature_b}
%\end{subfigure}
%\caption{(a) Landmarks $\boldsymbol{x}$ are generated randomly by the shape model in (\ref{eq:Model}) (\textit{blue dots}). (b) $d_1$ and $d_2$ are the SM feature values measuring the signed shortest distance from the myocardial boundary $B$ (\textit{blue contour}) to the pixel $\boldsymbol{p_1}$ and $\boldsymbol{p_2}$ respectively. $d_1$ is positive and $d_2$ is negative.}
%\label{fig:SMFeature}
%\end{figure}

\subsection{Shape Model Fitting} \label{sec:MethodFitting}
The RF probability map is an intermediate output that cannot be used directly for subsequent analysis and application. Simple post-processing on the probability map such as thresholding and edge detection produce noisy segmentations with false positives and incoherent boundaries due to the pixel-based nature of the RF classifier. In this section, we again make use of the shape model by fitting it to the RF probability map \cite{DBLP:conf/miccai/LiHCST16}. This generates a final closed myocardial contour that is smooth and coherent. The shape model allows only plausible myocardial shape which improves the segmentation accuracy by imposing shape constraints that correct for some of the misclassifications made by the RF.

Shape model fitting is formulated as an optimization problem where we want to find the optimal values of the shape and pose parameters $(\boldsymbol{b},\boldsymbol{\theta})$ such that the shape model best fit the RF probability map under some shape constraints. That is,
\begin{equation}\label{eq:fitting}
\begin{aligned}
& \underset{\boldsymbol{b},\boldsymbol{\theta}}{\text{min}}
& & { { \left\| { \boldsymbol{I} }_{ \textrm{RF} }-{ \boldsymbol{I} }_{ \textrm{M} }({ T }_{ \boldsymbol{\theta}  }(\bar { \boldsymbol{x} } +\boldsymbol{Pb})) \right\|  }^{ 2 }+\alpha \frac { 1 }{ K } \sum _{ i=1 }^{ K }{ \frac { \left| { b }_{ i } \right|  }{ \sqrt { { \lambda  }_{ i } }  }  }  } \\
& \text{subject to}
& & -{ s }\sqrt { { \lambda  }_{ i } } <{ b }_{ i }<{ s }\sqrt { { \lambda  }_{ i } }, \; i = 1, \ldots, K.
\end{aligned}
\end{equation}
The first term compares how well the model matches the RF probability map $\boldsymbol{I}_{\textrm{RF}}$. $\boldsymbol{I}_{\textrm{M}}(.)$ is a function which computes a binary mask from a set of landmarks generated by the shape model. This allows us to evaluate a dissimilarity metric between the RF probability map and the model binary mask by simply taking their sum-of-squared differences. The second term is a regularizer which imposes some shape constraints by keeping $b_i$'s small so that the final shape does not deviate too much away from the mean shape. This term is also related to the probability of the given shape \cite{DBLP:journals/pr/CristinacceC08}. $\alpha$ controls the weighting given to this regularization term. Another shape constraint is imposed on the objective function by setting the upper and lower bounds of $b_i$ to $\pm s\sqrt { { \lambda  }_{ i } }$ so that it can only vary within reasonable range similar to that of the shape model. The optimization is carried out using the pattern search algorithm \cite{doi:10.1137/S1052623400378742} since the objective function is not differentiable due to the difficulty of representing the derivative of the ${ \boldsymbol{I} }_{ \textrm{M} }$ function in mathematical form. The algorithm carries out global optimization which can be handled easily due to the small problem size (small number of shape and pose parameters). At the start of the optimization, we initialize each $b_i$ and $\theta_i$ to zero.

\subsection{Sequence Segmentation} \label{sec:MethodVidSeg}
We extend the above segmentation method for single 2D MCE image to 2D+t MCE sequence. The proposed method introduces temporal consistency to sequence segmentation by ensuring that the segmentation of the current frame does not differ too much from that of the previous frame. Specifically, an additional temporal constraint term is added to (\ref{eq:fitting}) in the shape model fitting step. The new objective function to minimize becomes 
\begin{equation}\label{eq:fitting2}
\begin{aligned}
& \underset{\boldsymbol{b},\boldsymbol{\theta}}{\text{min}}
& { { \left\| { \boldsymbol{I} }_{ \textrm{RF} }-{ \boldsymbol{I} }_{ \textrm{M} }({ T }_{ \boldsymbol{\theta}  }(\boldsymbol{x})) \right\|  }^{ 2 } +\alpha \frac { 1 }{ K } \sum _{ i=1 }^{ K }{ \frac { \left| { b }_{ i } \right|  }{ \sqrt { { \lambda  }_{ i } }  }  } }\\
& & {+\beta \frac { 1 }{ 2M } { \left\| \boldsymbol{x}-{ \boldsymbol{x} }_{ prev } \right\|  }^{ 2 } } \\
& \text{subject to}
& -{ s }\sqrt { { \lambda  }_{ i } } <{ b }_{ i }<{ s }\sqrt { { \lambda  }_{ i } }, \; i = 1, \ldots, K.
\end{aligned}
\end{equation}
where $\boldsymbol{x}=\bar { \boldsymbol{x} } + \boldsymbol{Pb}$ is the predicted landmark coordinates of the current frame and ${ \boldsymbol{x} }_{ prev }$ is the predicted landmark coordinates of the previous frame. The last term in (\ref{eq:fitting2}) is the temporal constraint which computes the sum-of-squared differences between the landmark positions of the two adjacent frames. The term makes use of the segmentation from the previous frame as a reference and penalizes any segmentation of the current frame which deviates too much away from the reference. The approach uses the previous segmentation as a guide for subsequent segmentation and this ensures the myocardial segmentations throughout the sequence transit smoothly in time. The temporal term is normalized by the number of landmarks $M$ and its influence is controlled by the weighting $\beta$. The segmentation of the previous frame is used as initialization for the current frame during optimization.

\section{Experiments} \label{sec:Experiments}
\subsection{Data and Annotations} \label{sec:ExperimentsData}
Our dataset consists of healthy subjects and CAD patients defined as those demonstrating $\geq$70\% luminal diameter stenosis of any major epicardial artery by qualitative coronary angiography. There are a total of 21 subjects of which 10 did not demonstrate CAD, 5 had single-vessel disease and 6 had multi-vessel disease. MCE exams from these 21 subjects were used in this paper. The data were acquired using a Philips iE33 ultrasound machine (Philips Medical Systems, Best, Netherlands) and SonoVue (Bracco Research SA, Geneva, Switzerland) as the contrast agent. For each of the 21 exams, MCE sequences in the apical 2, 3 and 4-chamber views are acquired. From these sequences, we randomly selected 242 2D MCE images to form $Dataset1$ which comprises an approximately equal proportion of the three different apical views and different cardiac phases (end-systole (ES) or end-diastole (ED)). $Dataset1$ is split into a training and a test set with a ratio of 14:7 on a patient basis and a ratio of 159:83 on an image basis. In addition, we chose 6 out of the 7 test subjects in $Dataset1$ and from these 6 subjects, we randomly chose 12 2D+t MCE sequences (2 for each subject) which again comprises an equal proportion of the three apical views. We then randomly select and crop out one cardiac cycle from each of these 12 sequences to form $Dataset2$. On average, each temporally cropped sequence in $Dataset2$ consists of 22$\pm$4 frames and captures one complete cardiac cycle from ES to ES.

Manual annotations for the two datasets are done by an expert. For $Dataset1$, each image is manually annotated with a bounding box which encloses the myocardium. This is used for training and testing the CNN. A myocardial contour is also manually delineated for each image in $Dataset1$ and for every image of the 12 sequences in $Dataset2$. This is used for training and testing the RF. Subsequently, 4 key landmarks are manually identified on the myocardial contour as illustrated in Fig. \ref{fig:TrainingShapes}. 18 landmarks are uniformly sampled in between each pair of key landmarks to give a total of $M$=76 landmarks that define each myocardial shape. The landmark annotations are used for shape model construction.

\subsection{Evaluation Metrics} \label{sec:ExperimentsMetrics}
Segmentation accuracy is evaluated quantitatively using Jaccard index which measures the overlap between two contours, mean absolute distance (MAD) which measures the average point-to-point distance between two contours and Hausdorff distance (HD) which measures the maximum distance between two contours. In addition, clinical indices such as endocardial and myocardial areas are also computed from the segmentations and compared to the ground truth in terms of correlation, bias and standard deviation. Correlation and Bland-Altman plots are also presented. Paired t-test is used to test for significant differences at 5\% significance level.

\subsection{Implementation Details} \label{sec:Implementation}
The CNN for the bounding box detection is implemented in Caffe \cite{DBLP:conf/mm/JiaSDKLGGD14} and run on a machine with one NVIDIA GeForce GTX 950 GPU. The parameters used for the CNN are described in Section \ref{sec:MethodBB}. 

The shape model is constructed with parameter $K$=16 $p$=99\% and $s$=2. The same shape model is used for both the SM feature and the shape model fitting. For the SM feature, we observe that the segmentation results are insensitive to $K$, $p$ and $s$. For the shape model fitting, these parameters have more significant influence on the results. Hence, we only tune their values for the shape model fitting and use the same values for the SM feature. To this end, only a single shape model needs to be constructed for the two tasks, making the approach more robust and generalizable. 

For the RF, we use 20 trees with a maximum tree depth of 24. Further increase in the number of trees in the forest and the depth of each tree add to the computational cost with no significant improvement in segmentation accuracy. All training images for the RF are pre-processed using histogram equalization to reduce intensity variations between different images. 10\% of the pixels from the training images are randomly selected for tree training. For the shape model fitting, we empirically set $\alpha$=3000 and $\beta$=10 via cross-validation.

Unless otherwise stated, 3-fold cross-validation is applied on the training set of $Dataset1$ to optimize the above parameters. Once the optimal parameters are found, the entire training set of $Dataset1$ is used to learn a CNN model, an RF model and a shape model. Using these models, testing is then performed on 1) the test set of $Dataset1$ and 2) the entire $Dataset2$.

Using a machine with Intel Core i7-4770 at 3.40 GHz and 32 GB of memory, RF training takes 7.1 minutes for 1 tree. Given a test image, RF segmentation takes about 25.5s using 20 trees but tree prediction can be parallelized so that it takes 1.3s per tree. The bounding box detection takes 0.1s and the shape model fitting takes 7.8s. In total, our fully automatic algorithm takes around 9.2s to segment one image.

\subsection{Comparisons with other State-of-the-Art Approaches} \label{sec:ExperimentsOther}
We compare our proposed approach to ASM for static segmentation on $Dataset1$ and to image registration and optical flow for sequence segmentation on $Dataset2$. 

We use a modified ASM \cite{DBLP:journals/tmi/GinnekenFSRV02} which selects a set of optimal features for the appearance model around each landmark point instead of using the Mahalanobis distance. The parameters for shape model construction are $K$=16 and $M$=76. The length of landmark intensity profile is 6 pixels. The search length for each landmark is 2 pixels and the number of search iterations is 40. The shape constraint is limited to $\pm 1.5\times \sqrt { { \lambda  }_{ i } } $ and two resolution levels are used for matching. The ASM is initialized by manually placing the contour near the myocardium.

The non-rigid image registration is based on B-spline free-form deformations  \cite{DBLP:journals/tmi/RueckertSHHLH99}. The error metric used is the sum of squared difference and two resolution levels are used. Smoothness penalty is set to 0.003 and B-spline control point spacing is set to 32 pixels. For each MCE sequence, the first frame is registered to all the other frames. A manual segmentation is performed on the first frame and it is propagated to the other frames in the sequence through the transformation field found by registration. 

Optical flow is based on the algorithm described in \cite{Brox2004}. It is a variational model based on the gray level constancy and the gradient constancy assumptions, together with a smoothness constraint. The weighting for the smoothness term is set to 0.05. A multi-resolution approach is also used with a downsampling ratio set to 0.75 and the width of the coarsest level set to 10 pixels. Optical flow motion fields are computed between consecutive frame pairs in the sequence and are used to propagate the manual segmentation on the first frame to all the other frames. 

The above methods all require some manual user inputs as initialization and the algorithm parameters are optimized using grid search.

\begin{figure*}
    \centering
    \begin{minipage}[b]{.26\linewidth}
        \centering
        \includegraphics[width=1.08\linewidth]{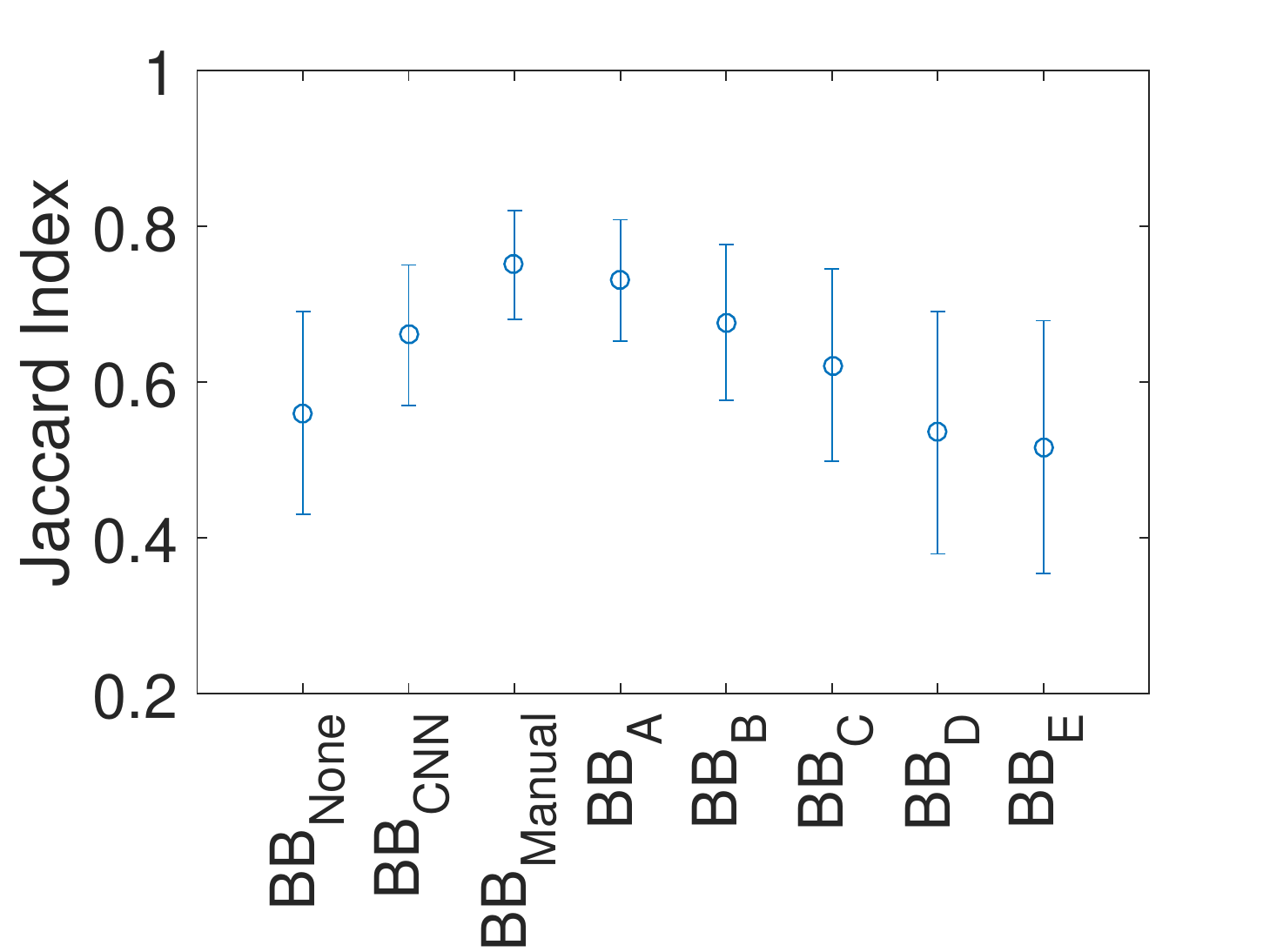}
    \end{minipage}%
    \hfill%
    \begin{minipage}[b]{.42\linewidth}
    	\centering
        \includegraphics[width=1\linewidth]{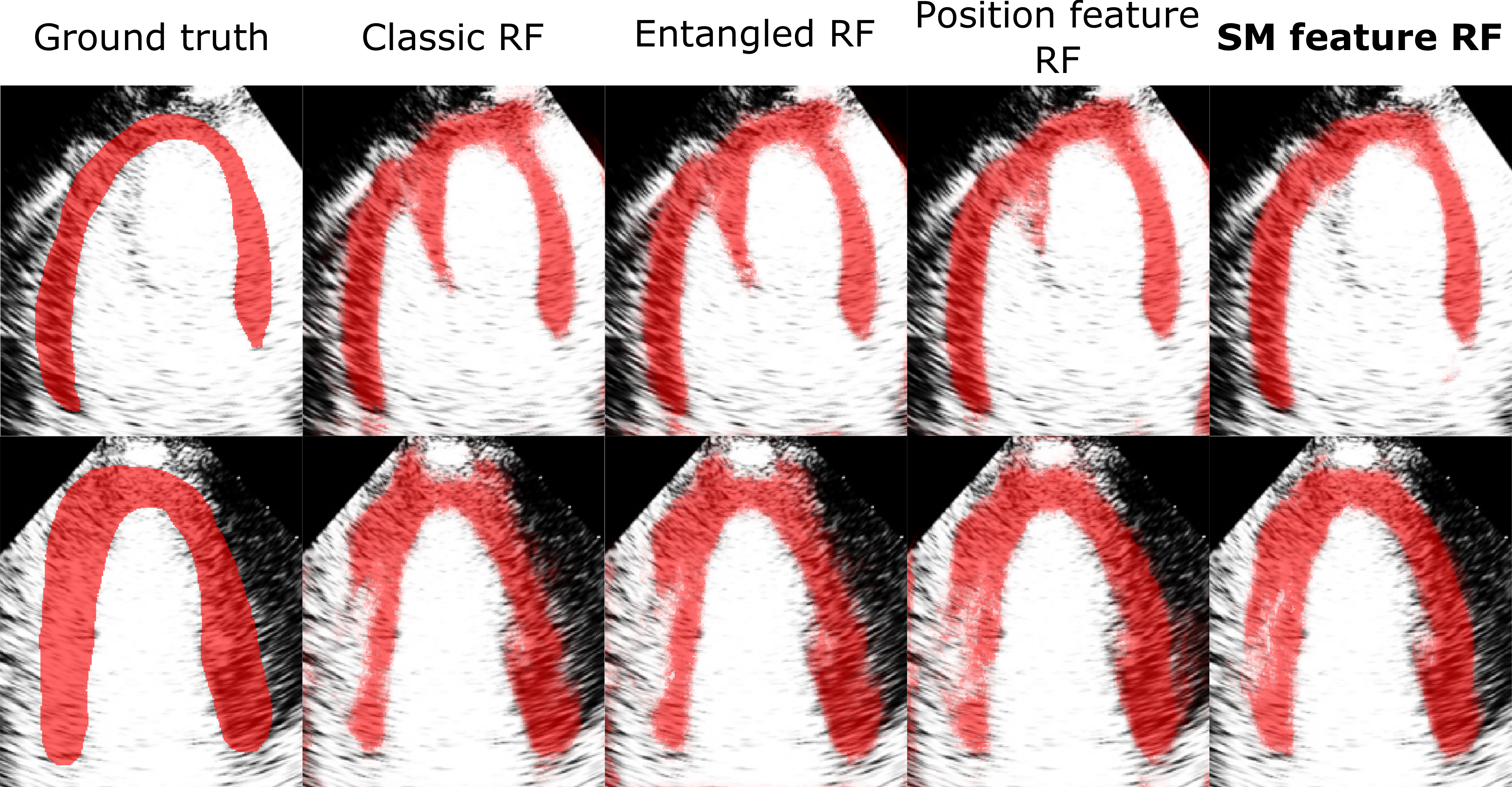}
    \end{minipage}%
    \hfill%
    \begin{minipage}[b]{.3\linewidth}
    	\centering
        \includegraphics[width=1\linewidth]{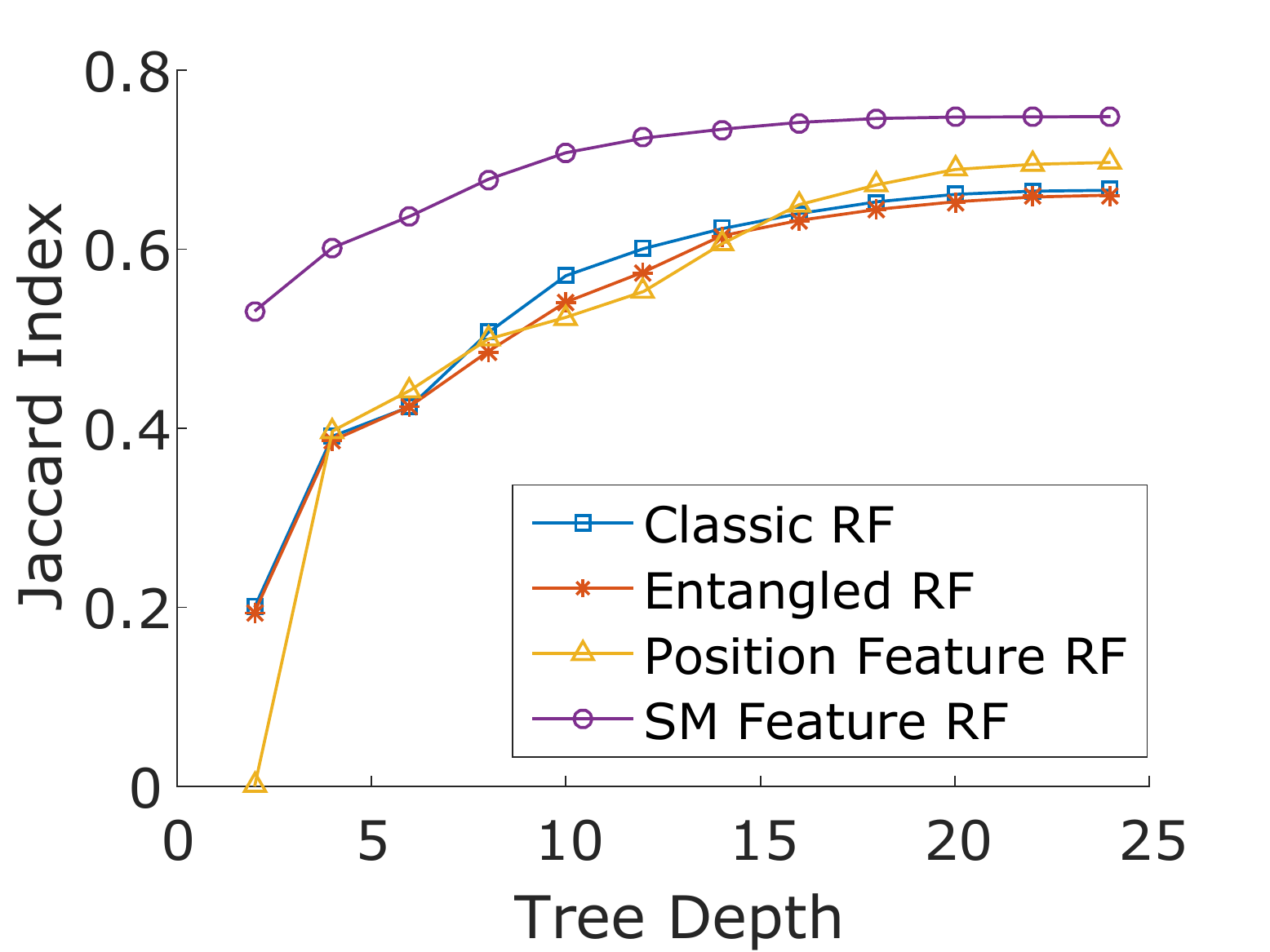}
    \end{minipage}\\[-7pt]
    \begin{minipage}[t]{.26\linewidth}
        \caption{Effect of bounding box detection on RF segmentation. Error bar denotes the standard deviation over the test set of $Dataset1$}
        \label{fig:BBEffect}
    \end{minipage}%
    \hfill%
    \begin{minipage}[t]{.42\linewidth}
        \caption{Visual comparison of the RF probability maps obtained by different RF variants.}
        \label{fig:ResultsVisual}
    \end{minipage}%
    \hfill%
    \begin{minipage}[t]{.3\linewidth}
        \caption{Comparison of Jaccard indices of different RF variants vs tree depths.}
        \label{fig:ResultsFeaturesDepth}
    \end{minipage}%
\end{figure*}

\section{Results} \label{sec:Results}
We evaluate different components of our segmentation method separately in the four sections below. Sections \ref{sec:ResultsBB} to \ref{sec:ResultsFinal} evaluate the bounding box detection, the SM feature and the proposed static segmentation algorithm respectively using $Dataset1$. Section \ref{sec:ResultsVidSeg} evaluates the sequence segmentation method using $Dataset2$.

\subsection{Bounding Box Detection} \label{sec:ResultsBB}
\begin{figure}
\centering
\includegraphics[width=\linewidth]{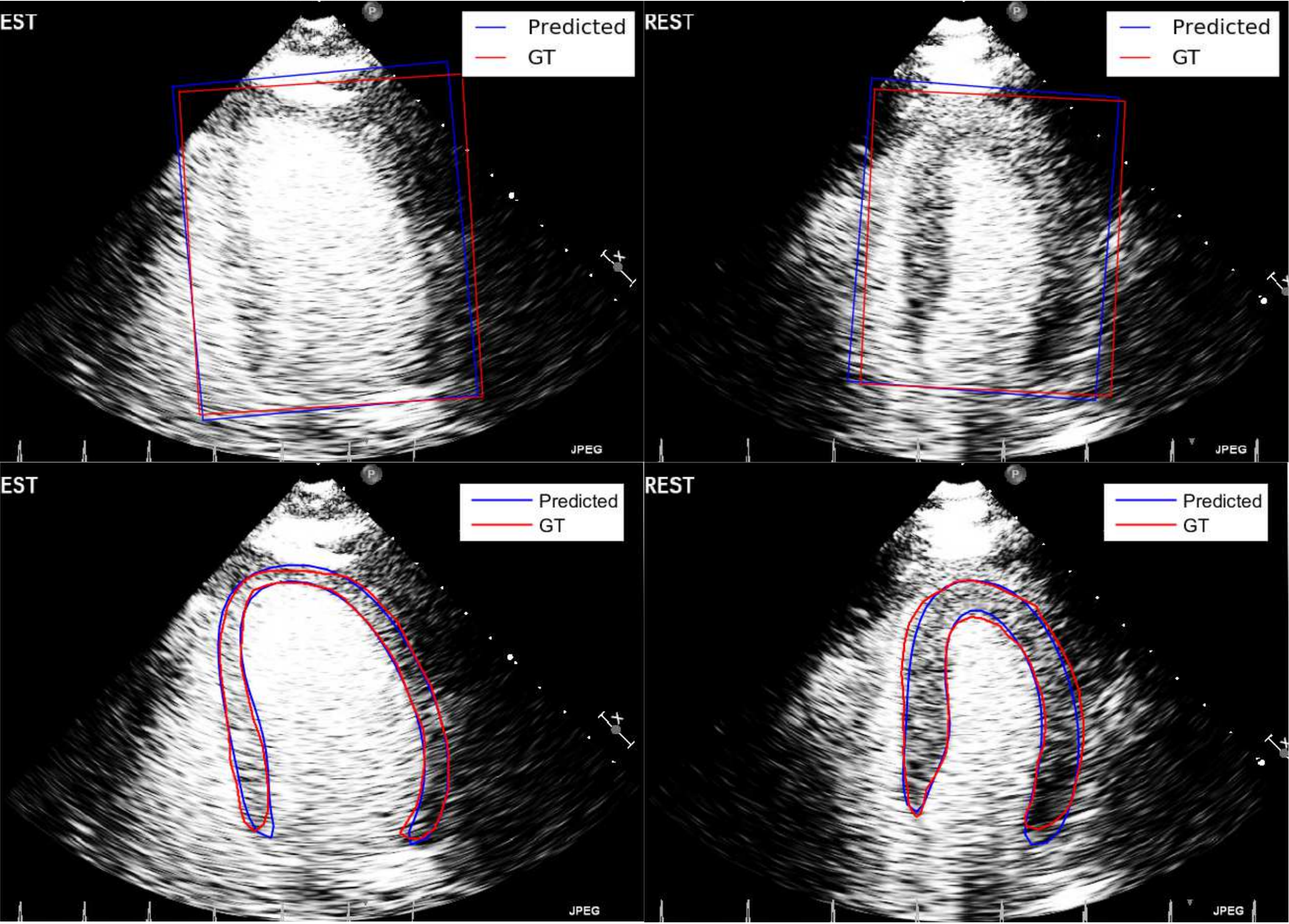}
\caption{Top: Visual comparison of the bounding boxes estimated by CNN (\textit{blue}) and the ground truth (\textit{red}). Bottom: Visual comparison of the final contours estimated by our fully automatic approach M2 (\textit{blue}) and the ground truth (\textit{red}).}
\label{fig:Bb_visual}
\end{figure}

In this experiment, we evaluate the accuracy of the CNN bounding box detection algorithm against the manual ground truth. We measure the detection error  as ${ { \boldsymbol{B} }_{ detected }-{ \boldsymbol{B} }_{ groundtruth }=\delta { \boldsymbol{B} }= }(\delta { B }_{ x }, \delta { B }_{ y }, \delta { B }_{ w }, \delta { B }_{ h }, \delta { B }_{ \theta  })$. Table \ref{table:ResultBBAccuracy} shows the bias and standard deviation of the error for each bounding box parameter. All parameters show mean values close to zero, indicating small systematic bias. Localization uncertainty of $({ B }_{ w }, { B }_{ h })$ is higher than that of $({ B }_{ x }, { B }_{ y })$. This indicates the CNN is less accurate in determining the scale of the bounding box compared to its position. CNN also has high localization precision for predicting the bounding box orientation $({ B }_{ \theta })$. Qualitatively, the top row of Fig. \ref{fig:Bb_visual} shows the manual and CNN-detected bounding boxes on some example MCE images.

In the next experiment, we study the effect of bounding box on the RF probability map. Specifically, we analyze three cases in which different input images are used for RF training and testing. \texorpdfstring{BB\textsubscript{None}}{BB None}: Original full size image; \texorpdfstring{BB\textsubscript{CNN}}{BB CNN}: Sub-image cropped from the bounding box detected by CNN; \texorpdfstring{BB\textsubscript{Manual}}{BB Manual}: Sub-image cropped from the manual bounding box; The RF probability maps obtained for these three cases are evaluated against the ground truth segmentations using Jaccard index as shown in Fig. \ref{fig:BBEffect}. Our SM feature works under the assumption that the shape of interest in all the RF input images are aligned to a reference. The bounding box effectively performs this alignment by cropping out a sub-image that is free from any myocardial pose variations which leads to more accurate RF segmentation. \texorpdfstring{BB\textsubscript{None}}{BB None} has the worst segmentation accuracy because it does not account for any pose variations. \texorpdfstring{BB\textsubscript{CNN}}{BB CNN} improves the segmentation accuracy but does not perform as well as \texorpdfstring{BB\textsubscript{Manual}}{BB Manual}. This is due to possible inaccuracy in the CNN-detected bounding box and pose variations may not be completely removed. However, \texorpdfstring{BB\textsubscript{CNN}}{BB CNN} is fully-automatic while \texorpdfstring{BB\textsubscript{Manual}}{BB Manual} requires the manual annotation of bounding box which makes the overall segmentation method semi-automatic. An additional advantage of the bounding box is that it removes irrelevant image regions and reduces the image size so that subsequent RF segmentation and shape model fitting is faster.

To further investigate the dependence of RF segmentation results on the bounding box detected, we conducted the following experiment. We perturb each ground truth bounding box ${\boldsymbol{B}_{groundtruth}}$ in the set \texorpdfstring{BB\textsubscript{Manual}}{BB Manual} by adding a random error $\delta { \boldsymbol{B} }$ to it. The error of each bounding box parameter $\delta { {B}_{i} }$, where $i=x,y,w,h,\theta$, is sampled randomly from a zero-mean normal distribution with standard deviation ${\sigma}_{{B}_{i}}$. By setting different ${\sigma}$ values, we create 5 sets of bounding boxes \texorpdfstring{BB\textsubscript{A}}{BB A}-\texorpdfstring{BB\textsubscript{E}}{BB E} with increasing amount of errors introduced to the ground truths. Specifically, we set ${{\sigma}_{\boldsymbol{B}}}=({ \sigma  }_{ { B }_{ x } }, { \sigma  }_{ { B }_{ y } }, { \sigma  }_{ { B }_{ w } }, { \sigma  }_{ { B }_{ h } }, { \sigma  }_{ { B }_{ \theta } })=(3.8, 2.9, 5.9, 6.5, 1.2)$ for \texorpdfstring{BB\textsubscript{A}}{BB A} and set multiples of it, $2{{\sigma}_{\boldsymbol{B}}},3{{\sigma}_{\boldsymbol{B}}},4{{\sigma}_{\boldsymbol{B}}},5{{\sigma}_{\boldsymbol{B}}}$, for \texorpdfstring{BB\textsubscript{B}}{BB B}, \texorpdfstring{BB\textsubscript{C}}{BB C}, \texorpdfstring{BB\textsubscript{D}}{BB D} and \texorpdfstring{BB\textsubscript{E}}{BB E}. $\sigma$ values of each bounding box parameter are chosen as such so that $\sigma$ values of \texorpdfstring{BB\textsubscript{C}}{BB C} are the same as the standard deviation of the CNN detection errors reported in Table \ref{table:ResultBBAccuracy}. Fig. \ref{fig:BBEffect} shows that RF segmentation accuracy decreases with increasing perturbations from the ground truth bounding boxes. This confirms that the final RF segmentation result is heavily dependent on the accuracy of bounding box detection. Future work should therefore be directed at improving the bounding box detection since there is still a significant gap for improvement on RF segmentations between using the CNN-detected bound boxes and the manual ground truths.

\def\arraystretch{1.2}
\begin{table}[]
\centering
\caption{Detection error of the bounding box parameters estimated by CNN against the manual ground truth. }
\label{table:ResultBBAccuracy}
\begin{tabular}{l l}
\hline \hline
Detection error & Mean $\pm$ SD \\
\hline
{ $\delta { B }_{ x }$ (pixels) } & 1.16$\pm$11.55 \\
{ $\delta { B }_{ y }$ (pixels) } & -2.47$\pm$8.74 \\
{ $\delta { B }_{ w }$ (pixels) } & 1.03$\pm$17.57 \\
{ $\delta { B }_{ h }$ (pixels) } & -1.28$\pm$19.58 \\
{ $\delta { B }_{ \theta }$ (\degree) } & 0.08$\pm$3.71 \\
\hline \hline
\end{tabular}
\end{table}

%\def\arraystretch{1.2}
%\begin{table}[]
%\centering
%\caption{$\sigma$ values of the Gaussian distributions used to randomly sampled the perturbation of each bounding box parameter. \texorpdfstring{BB\textsubscript{A}}{BB A}-\texorpdfstring{BB\textsubscript{E}}{BB E} are sets of bounding boxes with increasing perturbations from the ground truth \texorpdfstring{BB\textsubscript{Manual}}{BB Manual}. }
%\label{table:ResultBBPerturbations}
%\begin{tabular}{l l}
%\hline \hline
%& Perturbations $({ \sigma  }_{ { B }_{ x } }, { \sigma  }_{ { B }_{ y } }, { \sigma  }_{ { B }_{ w } }, { \sigma  }_{ { B }_{ h } }, { \sigma  }_{ { B }_{ \theta } })$ \\
%\hline
%\texorpdfstring{BB\textsubscript{Manual}}{BB Manual} & (0 0 0 0 0) \\
%\texorpdfstring{BB\textsubscript{A}}{BB A} & (3.8 2.9 5.9 6.5 1.2) \\
%\texorpdfstring{BB\textsubscript{B}}{BB B} & (7.7 5.8 11.7 13.1 2.5) \\
%\texorpdfstring{BB\textsubscript{C}}{BB C} & (11.5 8.7 17.6 19.6 3.7) \\
%\texorpdfstring{BB\textsubscript{D}}{BB D} & (15.3 11.6 23.5 26.1 4.9) \\
%\texorpdfstring{BB\textsubscript{E}}{BB E} & (19.2 14.5 29.3 32.7 6.2) \\
%\hline \hline
%\end{tabular}
%\end{table}

%\begin{figure}
%\centering
%\includegraphics[width=\linewidth]{BBEffect_big.eps}
%\caption{Effect of bounding box detection on RF segmentation.}
%\label{fig:ResultsBBEffect}
%\end{figure}

\begin{figure*}
\centering
\begin{subfigure}[b]{0.5\textwidth}
  \centering
  \includegraphics[width=0.95\linewidth]{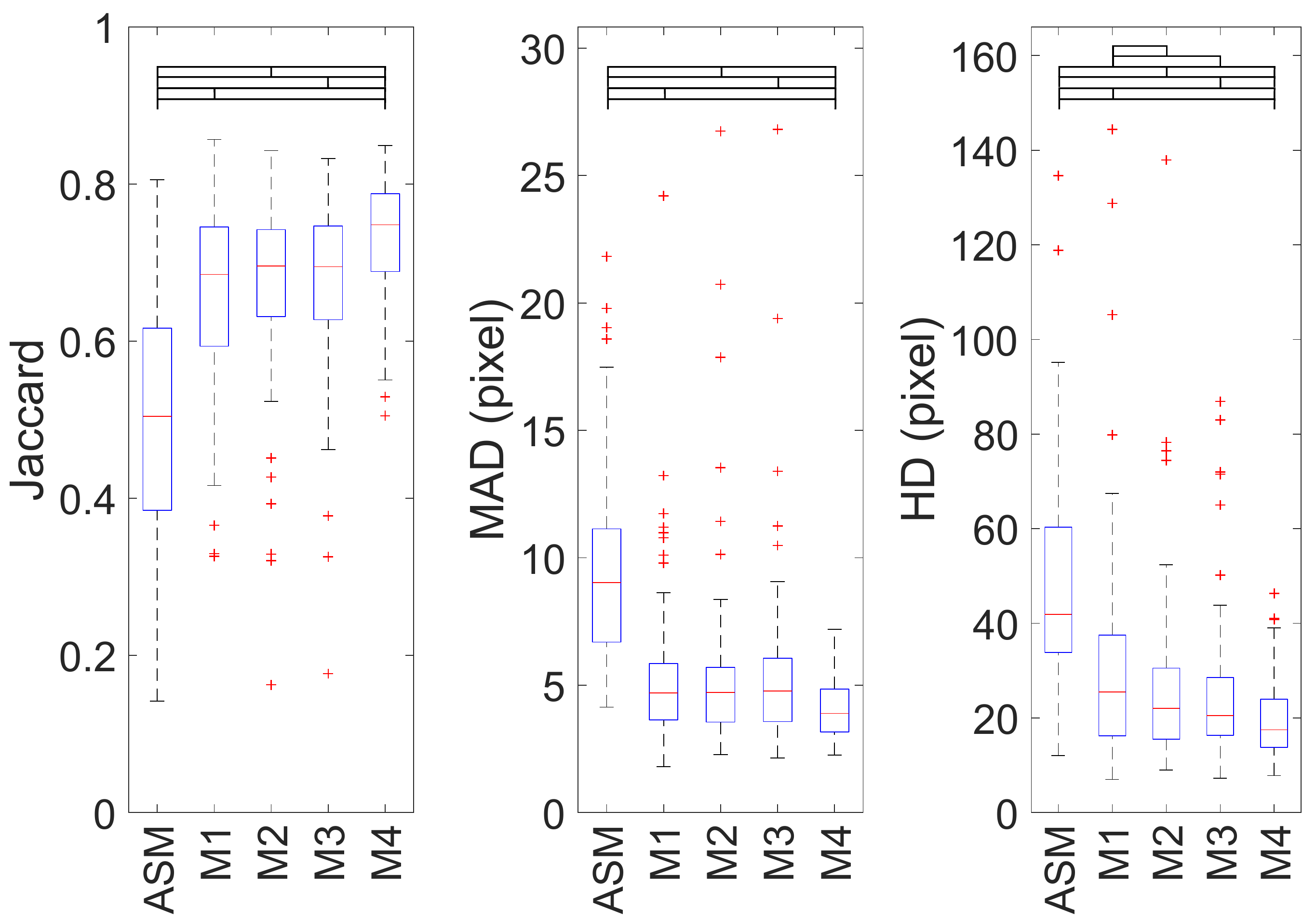}
  \caption{}
  \label{fig:ResultsSegAcc1}
\end{subfigure}%
\begin{subfigure}[b]{0.5\textwidth}
  \centering
  \includegraphics[width=0.95\linewidth]{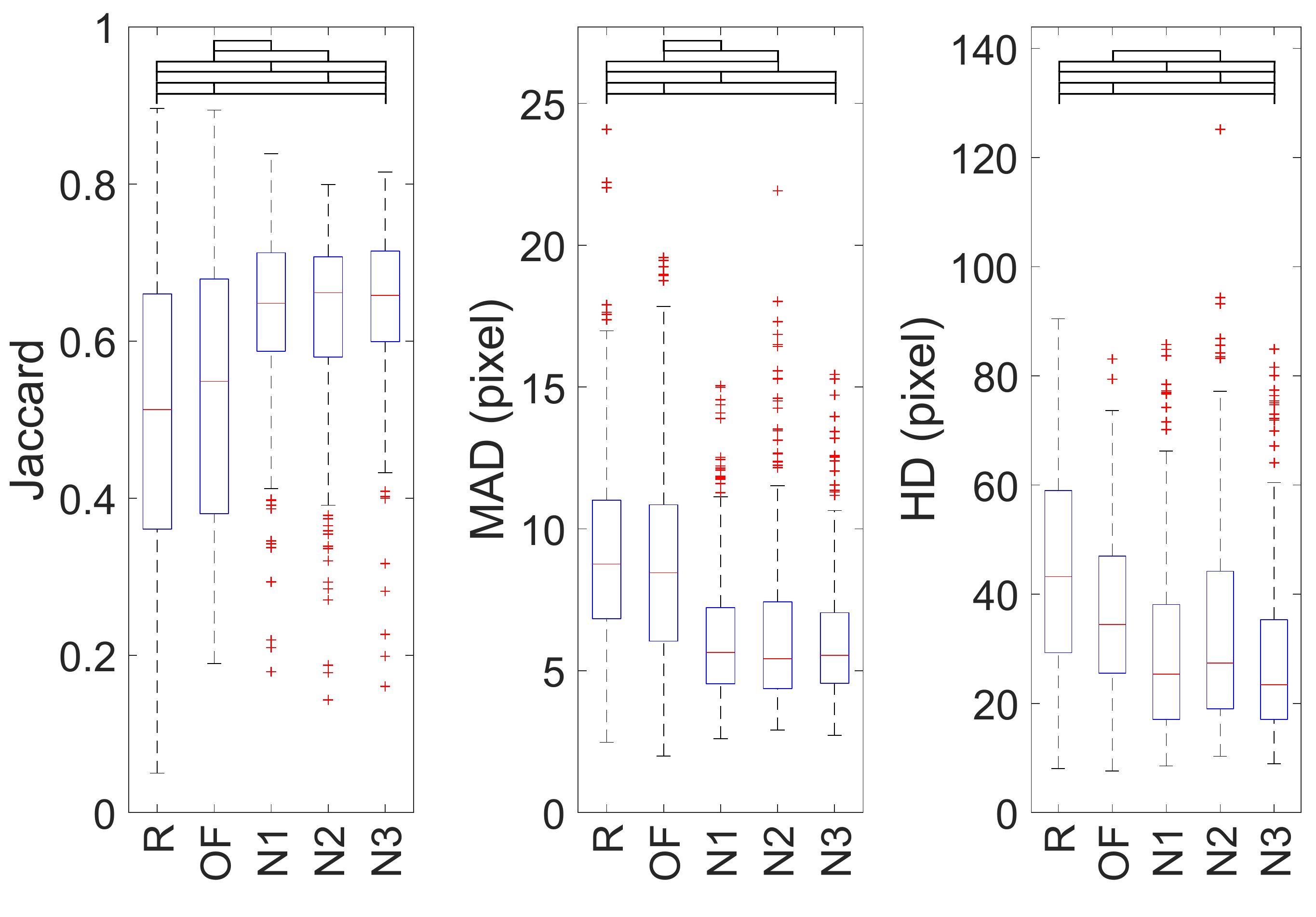}
  \caption{}
  \label{fig:ResultsSegAcc2}
\end{subfigure}
\caption{Segmentation accuracy results of the various approaches for (a) $Dataset1$ and (b) $Dataset2$. The ends of the whiskers represent the lowest and highest data point within 1.5 times the interquartile range. Top black brackets indicate the difference between the two approaches is significant using the paired t-test.}
\label{fig:ResultsSegAcc}
\end{figure*}

\subsection{Shape Model Feature} \label{sec:ResultsFeature}
We study the effect of SM feature on RF segmentation by comparing it with the classic RF that uses simple intensity features \cite{DBLP:journals/mia/CriminisiRKSPWS13} as well as RF that uses other contextual features such as the entanglement features \cite{DBLP:conf/ipmi/MontilloSWIMC11} and position features \cite{lempitsky_random_2009}. We use the sub-images extracted from the manual bounding boxes \texorpdfstring{BB\textsubscript{Manual}}{BB Manual} to train and test all the RFs in this section.

Fig. \ref{fig:ResultsVisual} compares the probability maps generated by RFs using different features. Our SM feature produces smoother and more coherent probability map. Other RFs often misclassify low intensity structure in the LV cavity as myocardium (row 1) and the high intensity region in the myocardium as background (row 2). RF with SM feature can overcome these problems by imposing a global shape constraint.

%\begin{figure*}
%\centering
%\begin{minipage}{.27\textwidth}
%  \centering
%  \includegraphics[width=1\linewidth]{BBEffect_big.eps}
%  \captionof{figure}{Effect of bounding box detection on RF segmentation.}
%  \label{fig:BBEffect_big}
%\end{minipage}%
%\begin{minipage}{.43\textwidth}
%  \centering
%  \includegraphics[width=1\linewidth]{rf_map_visual.pdf}
%  \captionof{figure}{Visual comparison of the RF probability maps obtained by different RF variants.}
%  \label{fig:ResultsVisual}
%\end{minipage}%
%\begin{minipage}{.3\textwidth}
%  \centering
%  \includegraphics[width=1\linewidth]{ContextFeaturesBigFont.pdf}
%  \captionof{figure}{Comparison of the Jaccard indices of different RF variants at different tree depths.}
%  \label{fig:ResultsFeaturesDepth}
%\end{minipage}
%\end{figure*}

%\begin{figure}
%\centering
%\includegraphics[width=\linewidth]{rf_map_visual.pdf}
%\caption{Visual comparison of the RF probability maps obtained using classic RF, entangled RF, position feature RF, SM feature RF.}
%\label{fig:ResultsVisual}
%\end{figure}

\def\arraystretch{1.2}
\begin{table}[]
\centering
\caption{Correlation coefficient, bias and standard deviation between ground truth and estimated clinical indices for $Dataset1$. Note that it is not possible to obtain clinical indices from the partial Canny edge maps of approach M1.}
\label{table:ResultInd1}
\begin{tabular}{l c c c c c c}
\hline \hline
& \multicolumn{3}{c}{Endocardial Area} & \multicolumn{3}{c}{Myocardial Area} \\
\cmidrule(lr){2-4}\cmidrule(lr){5-7}
& corr & bias & std & corr & bias & std \\
\cmidrule(lr){2-2}\cmidrule(lr){3-3}\cmidrule(lr){4-4}\cmidrule(lr){5-5}\cmidrule(lr){6-6}\cmidrule(lr){7-7}
%\hline
& (val) & \multicolumn{2}{c}{($\times$10\textsuperscript{3} pixel\textsuperscript{2})} & (val) & \multicolumn{2}{c}{($\times$10\textsuperscript{3} pixel\textsuperscript{2})} \\
\hline
ASM & 0.76 & -5.53 & 8.08 & 0.32 & 0.00 & 5.42 \\
M2 & 0.90 & -0.24 & 5.00 & 0.68 & -0.23 & 3.92 \\
M3 & 0.94 & -0.48 & 4.13 & 0.74 & -0.29 & 3.75 \\
M4 & 0.99 & -0.15 & 2.07 & 0.94 & 0.26 & 2.18 \\
\hline \hline
\end{tabular}
\end{table}

Fig. \ref{fig:ResultsFeaturesDepth} shows the segmentation performance of different RFs at different tree depths. Our SM feature RF outperforms the other RFs at all tree depths. This is because the SM feature captures the explicit geometry of the myocardium to guide the segmentation. A binary test of the SM feature in the split node partitions the image space using meaningful myocardial shape based on the shape model. Position feature \cite{lempitsky_random_2009} also learns the myocardial shape implicitly but its binary test partitions the image space using simple straight line which is less effective. To learn complex shape like the myocardium, many more position feature tests are needed compared to SM feature tests which learn the myocardial shape directly. This causes SM feature to be the more discriminative feature at lower levels of the tree, leading to better results than other RFs especially at lower tree depths.

%\begin{figure}
%\centering
%\includegraphics[width=0.8\linewidth]{ContextFeaturesBigFont.pdf}
%\caption{Quantitative comparison of the Jaccard performance of the classic RF, entangled RF, position feature RF, SM feature RF at different tree depths for 83 2D MCE test images from $Dataset1$.}
%\label{fig:ResultsFeaturesDepth}
%\end{figure}

\begin{figure*}
\centering
\includegraphics[width=1\linewidth]{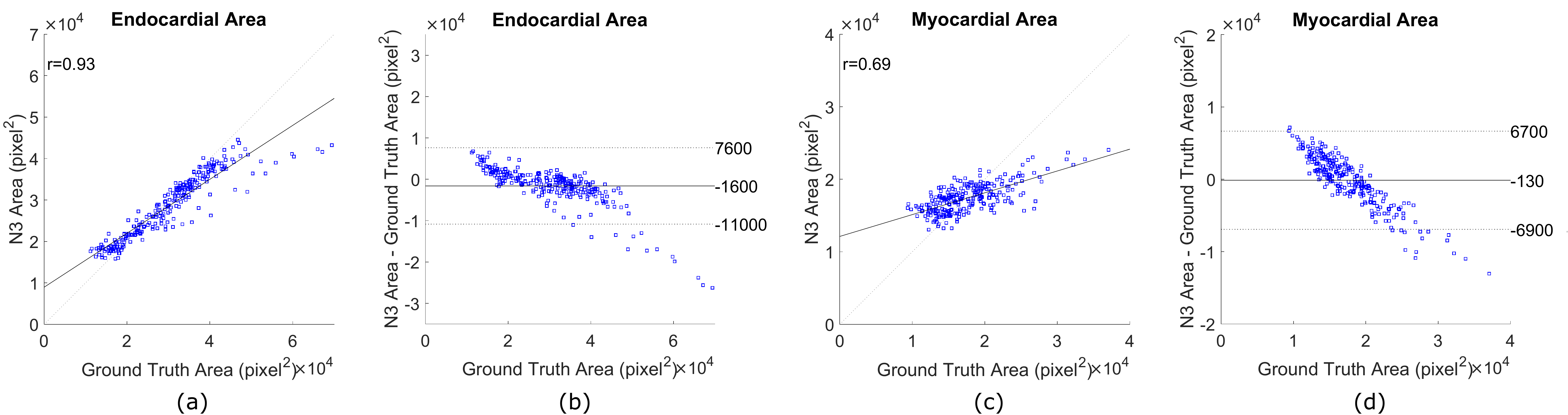}
\caption{Correlation plots (a,c) and Bland-Altman plots (b,d) of the endocardial and myocardial areas computed for the manual segmentations and estimated segmentations from N3. For BA plots, solid line: bias; dotted lines: limits of agreement $(\mu \pm 1.96 \sigma)$.}
\label{fig:ResultsInd2}
\end{figure*}

\subsection{Static Segmentation on $Dataset1$} \label{sec:ResultsFinal}
We compare the following different approaches for the evaluation of static segmentation on $Dataset1$.

\noindent ASM: Active shape Model \cite{DBLP:journals/tmi/GinnekenFSRV02}

\noindent M1: \texorpdfstring{BB\textsubscript{CNN}}{BB CNN} + SM feature RF + Canny edge detector

\noindent M2: \texorpdfstring{BB\textsubscript{CNN}}{BB CNN} + SM feature RF + Shape model fitting

\noindent M3: \texorpdfstring{BB\textsubscript{CNN}}{BB CNN} + SM feature RF + Shape model fitting (View specific)

\noindent M4: \texorpdfstring{BB\textsubscript{Manual}}{BB Manual} + SM feature RF + Shape model fitting  

Fig. \ref{fig:ResultsSegAcc1} presents the segmentation accuracy results (Jaccard, MAD and HD) and Table \ref{table:ResultInd1} reports the evaluation on clinical indices such as endocardial area and myocardial area. ASM does not perform well because it uses a simple intensity profile model to search for the best landmark position. This model is not adequate for noisy MCE images. RF can provide a much stronger and discriminative intensity model. In M1, we apply our segmentation method using the CNN bounding box detection algorithm (\texorpdfstring{BB\textsubscript{CNN}}{BB CNN}) and the RF with SM features. However, we replace the last shape model fitting step with a Canny edge detector to obtain a binary edge map as the final segmentation. This gives better results than the ASM but the final segmentation is not regularized by global shape constraint. To improve the results further, shape model fitting is added in M2 to give our fully automatic segmentation approach. It combines the local discriminative power of RF with the global shape constraint imposed by the shape model. The fitting guides the segmentation in regions where the RF probability map has low confidence predictions. It also ensures the final segmentation is a smooth and coherent contour that represents plausible myocardial shape. The bottom row of Fig. \ref{fig:Bb_visual} shows the final myocardial contour predicted by M2. The method is able to segment the myocardium accurately even in the presence of shadowing and attenuation artifacts which result in unclear epicardial border. In M3, we trained three separate RF models and shape models for the three different apical chamber views. There is a small improvement in results since each model learns more specifically the different anatomy of each view although a general model that includes all views (M2) is also quite robust. Since there are less training data for each view model in M3, we expect the results to improve with more data. In M4, we replace the CNN bounding box in M2 with manual bounding box (\texorpdfstring{BB\textsubscript{Manual}}{BB Manual}). This results in a semi-automatic approach which accurately removes any myocardial pose variations and achieves the best results in the final segmentation.

\subsection{Sequence Segmentation on $Dataset2$} \label{sec:ResultsVidSeg}
Fig. \ref{fig:ResultsSegAcc2} presents the segmentation accuracy of the different approaches on $Dataset2$. Our  proposed approach (N1-N3) achieves significantly more accurate segmentation results than image registration (R) and optical flow (OF) methods. Image registration and optical flow performs tracking by finding corresponding speckle patterns between frames and they are based on the constant intensity assumption. They do not perform well on MCE because MCE data exhibit high intensity variations and speckle patterns are decorrelated due to the highly dynamic microbubbles.

N1 is our static segmentation method which uses (\ref{eq:fitting}) for shape model fitting. It is the same as M2 and both RF model and shape model are trained on $Dataset1$ which consists of only ES and ED frames. N2 is exactly the same as N1 except that the training is performed on $Dataset2$ which consists of frames in all cardiac phases. Segmentation results for N2 are obtained using leave-one-out cross-validation on the 6 subjects in $Dataset2$. Although N2 is trained on all cardiac phases and is expected to have a more representative shape model, it is surprising to find that N1 has slightly more accurate results than N2. This could mean that the shape model trained from ED and ES frames alone is adequate for segmenting myocardial shapes over the full cardiac cycle. This is possible since the ED and ES shapes are at the extreme ends and all other shapes of the cardiac cycle can be found as intermediate transitions in between these two extremes. Another reason for the better performance of N1 could be due to N1 being trained on a more diverse dataset comprising 14 subjects while N2 is trained based on only 5 subjects during cross-validation. Increasing the number of training sequences from more subjects could potentially improve the results for N2. 

\begin{figure}
\centering
\includegraphics[width=0.8\linewidth]{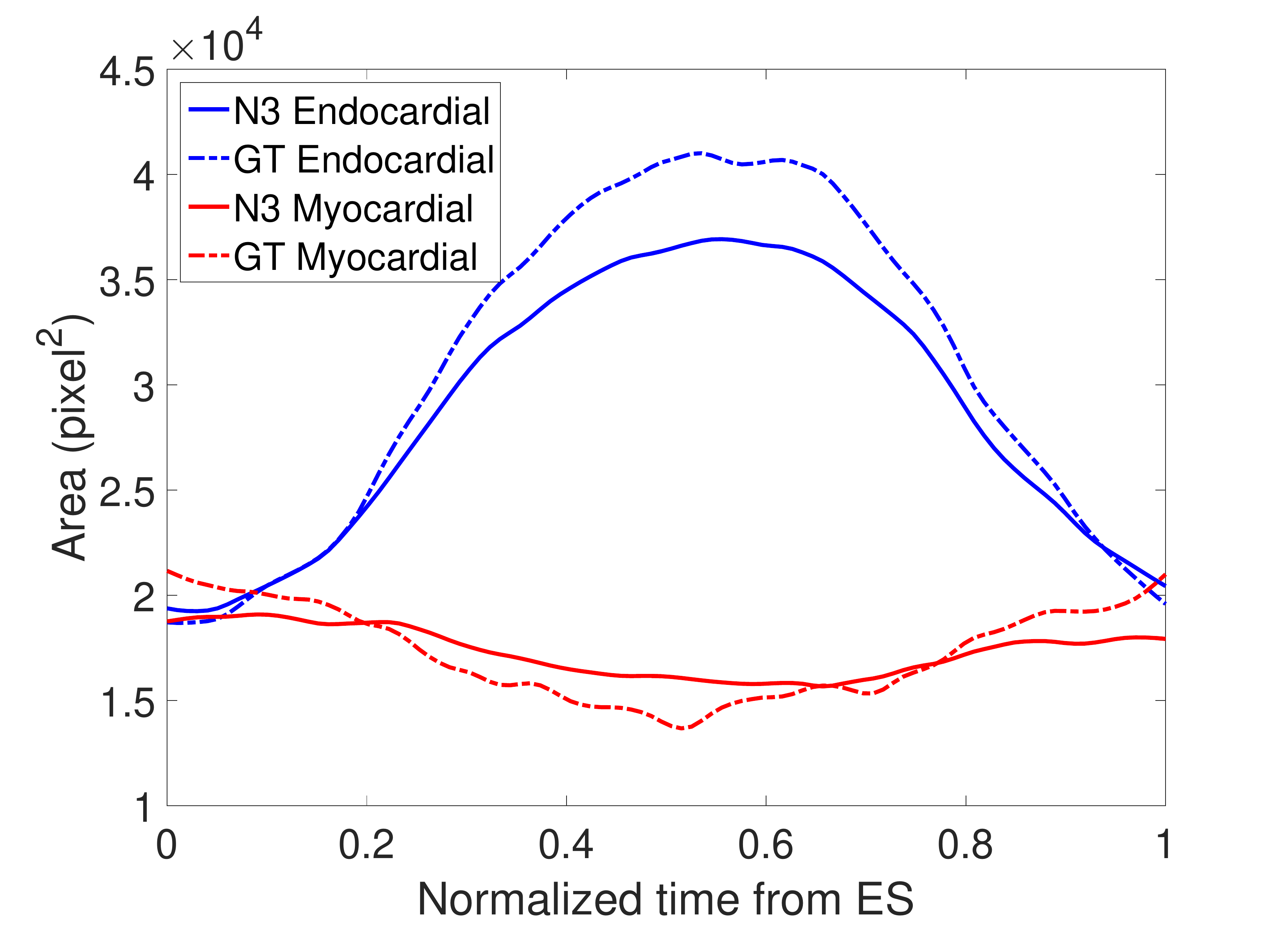}
\caption{Mean endocardial (blue) and myocardial (red) areas of 12 sequences over one cardiac cycle computed from ground truth (dash lines) and N3 segmentations (solid lines).}
\label{fig:ResultsAreaCurves}
\end{figure}

N3 is our sequence segmentation method and it achieves the best results by extending N1 using (\ref{eq:fitting2}) for shape model fitting. The temporal constraint allows N3 to produce temporally more consistent segmentations throughout the sequence with the most improvement reflected by the HD metric. The clinical indices of endocardial and myocardial areas are also computed for the automatic segmentations from N3 and compared to the manual segmentations in terms of correlation and Bland-Altman plots as shown in Fig. \ref{fig:ResultsInd2}. It can be observed that our method results in overestimation for smaller areas and underestimation for larger areas. Fig. \ref{fig:ResultsAreaCurves} also shows the myocardial and endocardial area-time curves over one cardiac cycle. The curves are the mean across the 12 sequences in $Dataset2$. We observe that our proposed method produces curves that are similar to the ground truth. Visual segmentation results on some full MCE sequences using N3 can be found in the supplementary materials.\footnote{Supplementary materials are available under the the "Supplementary Files" tab in ScholarOne Manuscripts.}

%\begin{figure}
%\centering
%\begin{subfigure}[b]{.24\textwidth}
%  \centering
%  \includegraphics[width=.8\linewidth]{VidSegUnrefineCrop.pdf}
%  %\includegraphics[height=.2\textheight]{Fig3}
%  \caption{}
%  \label{fig:VidSegUnrefine}
%\end{subfigure}%
%\begin{subfigure}[b]{.24\textwidth}
%  \centering
%  \includegraphics[width=.8\linewidth]{VidSegRefineCrop.pdf}
%  %\includegraphics[height=.2\textheight]{Fig4_matlab_bigFont}
%  \caption{}
%  \label{fig:VidSegRefine}
%\end{subfigure}
%\caption{Visual segmentation results of an 'outlier' frame. (a) Initially segmented using $Model_{Fit}$. (b) Re-segmented using $Model_{Vid2}$.}
%\label{fig:VidSeg}
%\end{figure}

\section{Discussion and Conclusion} \label{sec:Conclusion}
We have presented a fully automatic approach for fast and accurate segmentation of myocardium in 2D MCE image. The proposed method uses a statistical shape model to guide the RF segmentation by imposing global shape constraints. This is done by first incorporating a novel SM feature into the RF framework and then fitting the shape model to the RF probability map to obtain the final myocardial contour. The SM feature outperforms other contextual features by producing more accurate RF probability map while the shape model fitting step further improves the final segmentation results by producing a smooth and coherent contour. Bounding box detection using CNN serves as an important image alignment step that improves the performance of subsequent RF segmentation using SM feature. The method is further extended to 2D+t MCE sequence which imposes temporal consistency in the resultant sequence segmentations. The overall segmentation method combines the advantages of both ASM and RF, and outperforms either of this method used alone.

Our proposed method is generic and can be applied to other image data containing large intensity variations where prior knowledge of shape becomes important in guiding the segmentation. Current study only limits our approach to contrast echocardiography data. As future work, we will evaluate our method on other datasets of B-mode echocardiography or even medical data of different modalities in order to test the robustness and generality of the approach. In addition, our sequence dataset $Dataset2$ is small and only based on 6 subjects. Increasing the training data on this set can allow us to train better models than the one derived from $Dataset1$ which comprises only ES and ED frames. 

Extension to 3D data should also work in principle. In this case, we need to compute the shortest distance from a point to a surface for the SM feature which can increase computational cost. Shape model fitting can also take longer if 3D binary volume needs to be generated from the mesh surface of 3D shape model. Since this is the limiting step with the longest running time in our segmentation pipeline, future work will look at more efficient ways of optimizing this step in order to make our approach real-time. One way is to define the shape model on distance maps directly instead of the landmarks as described in \cite{1194625}. This will save the computational cost of converting the contour into a distance map or binary mask during SM feature computation and shape model fitting. 

We have added a simple temporal constraint term to induce temporally smooth sequence segmentation. More sophisticated tracking algorithms such as block matching and optical flow could be incorporated into our framework to ensure temporal consistency. This is done in \cite{Verhoek2011} where the RF segmentation is propagated using optical flow. Finally, effort should also be directed at more accurate bounding box detection since we have shown that the final segmentations obtained using our approach is heavily dependent on this step.

\section*{Acknowledgment}

This study was supported by the Imperial College President\textquotesingle s PhD Scholarships. The authors would like to thank Prof. Daniel Rueckert, Liang Chen and other members from the BioMedIA group for their help and advice.

\ifCLASSOPTIONcaptionsoff
  \newpage
\fi



\bibliographystyle{IEEEtran}
\bibliography{IEEEabrv,References_TMI2016}

%\begin{thebibliography}{1}
%\end{thebibliography}


%\end{IEEEbiography}

%\newpage



%\vfill


\end{document}